\def\year{2019}\relax
\documentclass[letterpaper]{article} 
\usepackage{aaai19}  
\usepackage{times}  
\usepackage{helvet}  
\usepackage{courier}  
\usepackage{url}  
\usepackage{graphicx}  
\usepackage{eucal}
\usepackage{amsmath}
\usepackage{enumerate}
\usepackage{tabularx}
\usepackage{hhline}
\usepackage{subcaption}
\usepackage{balance}
\usepackage{hyperref}

\newcommand{\model}{{FIA}~}
\newcommand{\modelns}{{FIA}}
\newcommand*{\defeq}{\stackrel{\text{def}}{=}}
\DeclareMathOperator*{\argmin}{arg\,min}

\nocopyright
\begin{document}
%
\title{Explaining Latent Factor Models for Recommendation with Influence Functions}
\author{Weiyu~Cheng, Yanyan~Shen\thanks{Corresponding author}, Yanmin~Zhu, Linpeng~Huang\\
Department of Computer Science and Engineering, Shanghai Jiao Tong University\\
Email: \{weiyu\_cheng, shenyy, yzhu, lphuang\}@sjtu.edu.cn\\
}
\maketitle
\begin{abstract}
 
Latent factor models (LFMs) such as matrix factorization achieve the state-of-the-art performance among various Collaborative Filtering (CF) approaches for recommendation. Despite the high recommendation accuracy of LFMs, a critical issue to be resolved is the lack of explainability. Extensive efforts have been made in the literature to incorporate explainability into LFMs. However, they either rely on auxiliary information which may not be available in practice, or fail to provide easy-to-understand explanations. In this paper, we propose a fast influence analysis method named FIA, which successfully enforces explicit neighbor-style explanations to LFMs with the technique of influence functions stemmed from robust statistics. We first describe how to employ influence functions to LFMs to deliver neighbor-style explanations. Then we develop a novel influence computation algorithm for matrix factorization with high efficiency. We further extend it to the more general neural collaborative filtering and introduce an approximation algorithm to accelerate influence analysis over neural network models. Experimental results on real datasets demonstrate the correctness, efficiency and usefulness of our proposed method.

\end{abstract}

\section{Introduction}
Recommender systems play an increasingly significant role in improving user satisfaction and revenue of content providers who offer personalized services. Collaborative filtering (CF) methods, aiming at predicting users' personalized preferences against items based on historical user-item interactions, are the primary techniques used in modern recommender systems. Among various CF methods, latent factor models (LFMs) such as matrix factorization (MF), have gained popularity via the Netflix Prize contest~\cite{DBLP:journals/sigkdd/BellK07} and achieved the state-of-the-art performance.


The key idea of LFMs is to learn latent vectors for users and items in a low-dimensional space, and each user-item preference score is typically modeled as a function of two (i.e., user and item) latent vectors, e.g., performing simple inner product, or non-linear transformation with neural structures~\cite{DBLP:conf/www/HeLZNHC17}. In spite of the superior performance, a critical issue with LFMs to be resolved is the lack of explainability. To be specific, it is extremely difficult to interpret each latent dimension and explain why the preference scores are derived in a particular manner.  
On the contrary, most neighbor-based CF models, which generally perform worse than LFMs, are explainable thanks to their inherent algorithm design. For example, item-based CF~\cite{DBLP:conf/www/SarwarKKR01} recommends an item to a user by telling ``this item is similar to some of your previously liked items'', while user-based CF~\cite{DBLP:conf/cscw/ResnickISBR94} explains the recommendation of an item by saying ``several users who are similar to you liked this item''. We dub such intuitive explanations as \emph{neighbor-style explanations}. 
Enforcing explainability in recommender systems is inevitably important, which can make the reasoning more transparent and improve the trustworthiness and users' acceptance of the recommendation results~\cite{bilgic2005explaining,DBLP:reference/rsh/2011}.
In this paper, we aim to answer the following question: can we endue latent factor models with the ability of providing neighbor-style explanations?


Extensive efforts have been made in the literature to incorporate explainability into LFMs, which mainly fall into two categories: \emph{content-based settings} and \emph{collaborative settings}~\cite{DBLP:journals/corr/abs-1804-11192}. 
For content-based settings, many researches focused on extracting explicit item features from auxiliary information to express the semantics of latent dimensions.
For example,~\citeauthor{DBLP:conf/sigir/ZhangL0ZLM14}~\shortcite{DBLP:conf/sigir/ZhangL0ZLM14} proposed the Explicit Factor Model, which extracts product features based on user reviews and aligns each latent factor dimension with a product feature towards explainable recommendation. ~\citeauthor{DBLP:conf/sigir/WangWJY18}~\shortcite{DBLP:conf/sigir/WangWJY18} and~\citeauthor{DBLP:conf/www/ChengDZK18}~\shortcite{DBLP:conf/www/ChengDZK18} empowered LFMs with explanations using both product features and review aspects. However, the external information (e.g. user reviews) required by the existing content-based filtering methods may not be available in practice. And the latent space composed from the recognized explicit item features (usually less than 10 understandable features) could be insufficient to represent a large number of users and items (up to hundred millions) while preserving their semantic similarities and dissimilarities.

As for the collaborative settings, non-negative matrix factorization was proposed to enhance the interpretability of MF methods by adding a non-negative constraint on the factors~\cite{DBLP:conf/nips/LeeS00}, but it fails to provide explicit explanations. 
More recently, ~\citeauthor{DBLP:conf/recsys/AbdollahiN17}~\shortcite{DBLP:conf/www/AbdollahiN16,DBLP:conf/recsys/AbdollahiN17} introduced Explainable Matrix Factorization which prefers to recommend items liked by user's neighbors based on an ``explainability regularizer''. A similar idea is also applied to the restricted Boltzmann machines for CF~\cite{DBLP:journals/corr/AbdollahiN16}. But the purpose of these two works is to improve some ad-hoc ``explainability scores'' (e.g., the number of a user's neighbors who liked the recommended item) rather than generate explanations for LFMs.
Furthermore, the above collaborative methods require specific modifications to the vanilla LFMs, and hence their explanation abilities can hardly be applicable in the LFM variants, e.g., Neural Collaborative Filtering (NCF)~\cite{DBLP:conf/www/HeLZNHC17}.
In this paper, we propose a general method named \model ({\underline F}ast {\underline I}nfluence {\underline A}nalysis), which successfully enforces neighbor-style explanations to LFMs 
and only relies on the user-item rating matrix without the auxiliary information requirement.
The key technique used in \model is the \emph{influence functions} stemmed from robust statistics~\cite{cook1980characterizations}.
Influence functions were originally developed to understand the influence of training examples on a model's predictions~\cite{cook1982residuals}. 
In the context of recommendation, we train an LFM using users' historical item ratings and the trained model can predict ratings for the unrated items.
Given a trained LFM and its predicted rating for a specific user-item pair, influence functions allow us to identify \emph{which training example, in the form of (user, item, rating), contributes most to the prediction result}. By aligning the user and item properly, we 
are able to evaluate the effects of the historical ratings from the same user (or, from the same item) on the rating predicted by the model. Naturally, the historical ratings with the maximum influence form a neighbor-style explanation for the LFM towards explainable recommendation. Note that the enforcement of influence functions is orthogonal to the specific LFM structures and can be seamlessly applied to different LFMs.

To the best of our knowledge, this is the first attempt to leverage influence functions in the recommendation domain to facilitate neighbor-style explanations in LFMs. 
The key technical challenge of employing influence functions in LFM-based recommendation is the high computation cost, which is determined by the large number of model parameters in advanced LFMs and the scale of training data.
To make influence analysis applicable, our proposed \model introduces an efficient influence calculation algorithm which exploits the characteristics of MF to effectively reduce the time complexity. We further extend our algorithm to the more general neural LFM, i.e., NCF, and develop an approximation algorithm to accelerate influence analysis over neural methods.
Extensive experiments have been conducted over real-world datasets and the results demonstrate the correctness and efficiency of \modelns. The analysis on the results of influence functions leads to better understanding of LFM behaviors, which is valuable for a broader domain of recommendation-related researches.


\section{Preliminaries}

\subsection{Latent Factor Models}

Matrix Factorization (MF) has become the \emph{de facto} approach to LFM-based recommendation. MF represents each user/item with a real-valued vector of latent features. Let $\mathbf{p}_u$ and $\mathbf{p}_i$ denote the vectors for user $u$ and item $i$ in a joint $K$-dimensional latent space, respectively. In MF, the predicted rating $\hat{R}_{ui}$ of user $u$ against item $i$ is computed by the inner product of $\mathbf{p}_u$ and $\mathbf{q}_i$, as defined below:
\begin{equation}
\label{equ:basic}
		\hat{R}_{ui}=\mathbf{p}_u^T\mathbf{q}_i
\end{equation}
The inner product operation linearly aggregates the pairwise latent feature multiplications, which limits the expressiveness of MF in capturing complex user-item interactions. Neural Collaborative Filtering (NCF)~\cite{DBLP:conf/www/HeLZNHC17} is thus proposed to learn a non-linear \emph{interaction function} $f(\cdot)$, which can be considered as a generalization of MF:
\begin{equation}
	\hat{R}_{ui}=f(\mathbf{p}_u,\mathbf{q}_i)
\end{equation}
In the original paper, $f(\cdot)$ is specialized by a multi-layer perceptron (MLP):
\begin{equation}
	\hat{R}_{ui}=\phi_X(...\phi_2(\phi_1(\mathbf{z}_0))...)
\end{equation}
\begin{equation}
	\mathbf{z}_0= \mathbf{p}_u\oplus\mathbf{q}_i
\end{equation}
\begin{equation}
	\phi_l(\mathbf{z}_{l-1}) = \delta_l(\mathbf{W}_l\mathbf{z}_{l-1}+\mathbf{b}_l),~~l\in[1,X]
\end{equation}
where $X$ is the number of fully-connected hidden layers in the neural network, $\oplus$ is the concatenation of vectors, $\mathbf{W}_l$, $\mathbf{b}_l$ and $\delta_l$ are the weight matrix, bias vector and non-linear activation function for the $l$-th layer, respectively. Several recent attempts~\cite{DBLP:conf/ijcai/0001DWTTC18,DBLP:conf/ijcai/ChengSZH18} replace MLP with more complex operations (e.g., convolutions), but they still belong to the general framework of NCF. In this paper, we mainly focus on MF and the original NCF method for clarification, but our proposed algorithms are applicable to all the instantiated models under the NCF framework.


\subsection{Influence Functions}
\label{sec:pre:if}

Consider a general prediction problem from an input domain $\CMcal{X}$ to an output domain $\CMcal{Y}$. Let $Z=\{z_1, z_2, ..., z_n\}$ be the training set, where $z_i=(x_i,y_i) \in \CMcal{X} \times \CMcal{Y}, \forall i\in\{1, n\}$. Given a point $z\in Z$ and model parameters $\theta \in \Theta$, we denote by $L(z, \theta)$ the empirical loss on $z$. The objective function $R(\theta)$ for model training is defined as $\frac{1}{n}\sum_{i=1}^n{}L(z_i, \theta)$, and the model parameters that minimize $R(\theta)$ is defined as $\hat{\theta} \defeq \argmin_{\theta \in \Theta} \frac{1}{n}\sum_{i=1}^nL(z_i, \theta)$. We assume $R(\theta)$ is twice-differentiable and strictly convex in $\theta$, and this assumption can be relaxed via approximation.

The ultimate goal of using influence functions is to estimate the effects of training points on a model's predictions. A simple solution to achieve this goal is to remove a training point, retrain the model with the remaining points from scratch, and compute the new prediction results. However, this process involves high time consumption. Influence functions provides an efficient way to estimate the model's prediction change without retraining the model.
%
This is achieved by studying the change of model parameters $\hat{\theta}_{\epsilon,z} - \hat{\theta}$ when a training point $z$ is upweighted by an infinitesimal step $\epsilon$, and the new parameters $\hat{\theta}_{\epsilon,z} \defeq \argmin_{\theta \in \Theta} \frac{1}{n}\sum_{i=1}^nL(z_i, \theta) + \epsilon L(z,\theta)$. A classical result~\cite{cook1982residuals} claims that:
\begin{equation}
	\frac{{\rm d} \hat{\theta}_{\epsilon,z}}{{\rm d} \epsilon}\Big|_{\epsilon=0}=-H_{\hat{\theta}}^{-1}\nabla_{\theta}L(z,\hat{\theta})\label{eq:dtepsilon}
\end{equation}
where $H_{\hat{\theta}} \defeq \frac{1}{n}\sum_{i=1}^n \nabla_{\theta}^2 L(z_i, \hat{\theta})$ is the Hessian matrix. Equation~(\ref{eq:dtepsilon}) is derived from a quadratic approximation to $R(\theta)$ via Taylor expansion. $H_{\hat{\theta}}$ is invertible by the assumption on $R(\theta)$. Recent works have shown that for non-convex objective functions that are widely used in neural networks, a damping term can be added into the Hessian to alleviate negative eigenvalues and make the equation approximately work.
The detailed derivations can be found in~\cite{DBLP:conf/icml/KohL17}.

We then measure the change on loss at a test point $z_{test}$ if upweighting $z$ by the step $\epsilon$ using the chain rule:
\begin{equation}
\label{eq:dztest}
\begin{aligned}
	\frac{{\rm d} L(z_{test},\hat{\theta}_{\epsilon,z})}{{\rm d} \epsilon}\Big|_{\epsilon=0}
	&=\nabla_{\theta}L(z_{test},\hat{\theta})^\top \frac{{\rm d} \hat{\theta}_{\epsilon,z}}{{\rm d} \epsilon}\Big|_{\epsilon=0}\\
	&=-\nabla_{\theta}L(z_{test},\hat{\theta})^\top H_{\hat{\theta}}^{-1}\nabla_{\theta}L(z,\hat{\theta})
\end{aligned}
\end{equation}

By setting $\epsilon$ to $-\frac{1}{n}$ (which is equivalent to removing the point $z$), we can approximate the influence of removing $z$ from the training set on the loss at the test point $z_{test}$:
\begin{equation}
\small
\label{eq:testloss}
\begin{aligned}
L(z_{test},\hat{\theta}_{-z}) 
&\approx L(z_{test},\hat{\theta})-\frac{1}{n}\frac{{\rm d} L(z_{test},\hat{\theta}_{\epsilon,z})}{{\rm d} \epsilon}\Big|_{\epsilon=0}\\
&=L(z_{test},\hat{\theta})+\frac{1}{n}\nabla_{\theta}L(z_{test},\hat{\theta})^\top H_{\hat{\theta}}^{-1}\nabla_{\theta}L(z,\hat{\theta})
\end{aligned}
\normalsize
\end{equation}
where $\hat{\theta}_{-z}$ is the learned model parameters after removing $z$. From the above equation, we can infer the influence of a training point $z$ on model's prediction loss at $z_{test}$ with the previously learned model parameters $\hat{\theta}$ without retraining.

\section{Methodology}

In this section, we first describe how to apply influence functions to LFMs to deliver neighbor-style explanations. We then introduce our fast influence analysis method \model for MF that significantly reduces the computation cost. Finally, we extend our method to neural settings and propose an approximation algorithm to further improve analysis efficiency over NCF.

\subsection{Explaining LFMs via Influence Analysis}

Consider the rating prediction problem for recommendation where the input space $\CMcal{X}$ involves the sets of users $\{u_j\}$ and items $\{i_j\}$, and the output space $\CMcal{Y}$ is the set of possible ratings. 
Let $\CMcal{R} = \{z_1, z_2, ..., z_n\}$ be a set of observed user-item ratings, where $z_j=(u_j,i_j,y_j) \in \CMcal{X} \times \CMcal{Y}$. Without loss of generality, we define $L(z_j, \theta)$ as the squared error loss at ($u_j, i_j$) reported by an LFM with parameters $\theta$:
\begin{equation}
	L(z_j, \theta) = (g(u_j,i_j,\theta) - y_j)^2
\end{equation}
where $g(u_j,i_j,\theta)$ is the model's predicted rating for $(u_j,i_j)$. The LFM is trained based on $\CMcal{R}$ and the model parameters $\hat{\theta}$ on convergence satisfies: $\hat{\theta} \defeq \argmin_{\theta} \frac{1}{n}\sum_{j=1}^nL(z_j, \theta)$.

\begin{figure}[t!]
	\centering
	\subcaptionbox{Item-based explanation\label{subfig:item-based}}
	[.49\linewidth]{\includegraphics[width=.49\linewidth]{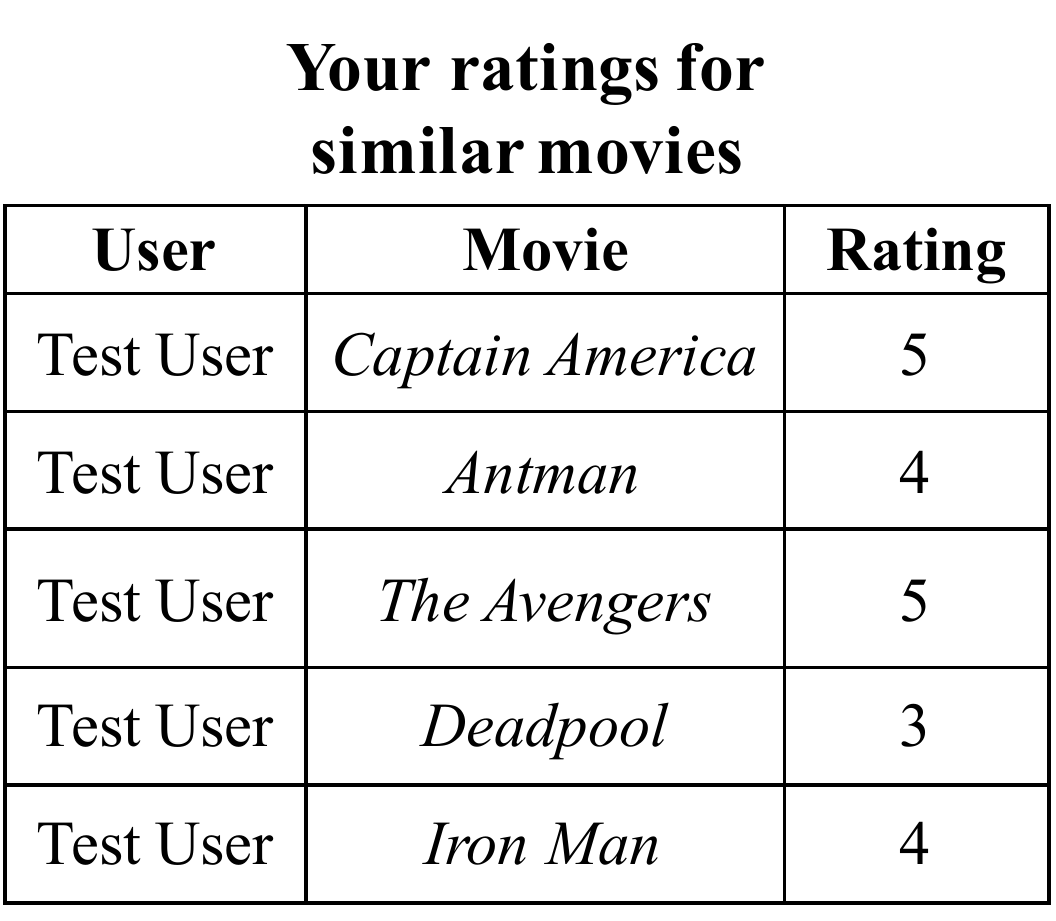}}
	\subcaptionbox{User-based explanation\label{subfig:user-based}}
	[.49\linewidth]{\includegraphics[width=.49\linewidth]{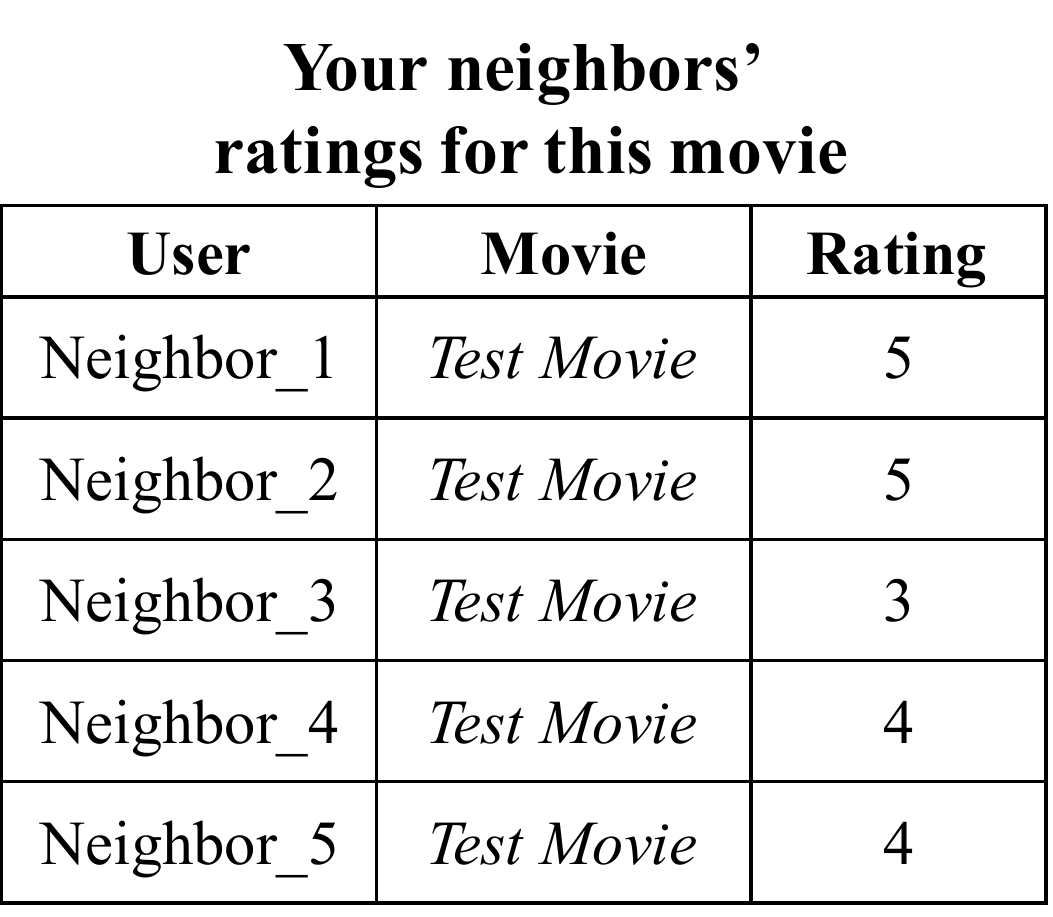}}
	\caption{Two types of neighbor-style explanations.}\label{fig:explain}
\end{figure}

\noindent {\bf Problem.}
We denote by $g(u_{t},i_{t},\hat{\theta})$ the rating predicted by the trained LFM over a test case $(u_{t},i_{t})$.  
Let $\CMcal{R}_{u_t}$ and $\CMcal{R}_{i_t}$ be two subsets of training records that are interacted with $u_t$ and $i_t$, respectively. That is:
$${\CMcal R}_{u_t}=\{z_j=(u_j, i_j, y_j)\in \CMcal{R} \mid u_j=u_t \}$$
$${\CMcal R}_{i_t}=\{z_j=(u_j, i_j, y_j)\in \CMcal{R} \mid i_j=i_t \}$$
In this paper, we aim to find the top-$k$ influential points in $\CMcal{R}_{u_t}$ that lead to model's prediction $g(u_{t},i_{t},\hat{\theta})$. Similarly, we also aim to find the $k$ most influential points in $\CMcal{R}_{i_t}$ that make the model predict $g(u_{t},i_{t},\hat{\theta})$. We dub the above two sets of identified influential points as \emph{item-based} and \emph{user-based} neighbor-style explanations respectively, as illustrated in Figure~\ref{fig:explain}, where $k=5$. Note that the influence analysis process should be efficient and scalable to a large number of users and items.
 

\noindent {\bf Influence analysis towards explainable LFMs.}
Inspired by the power of influence functions, for a training point $z \in \CMcal{R}_t$, where we define $\CMcal{R}_t \defeq \CMcal{R}_{u_t}\!\cup\!\CMcal{R}_{i_t}$, we can measure its influence on a prediction $g(u_{t},i_{t},\hat{\theta})$ by studying the counter-factual: \emph{How would this prediction change if we did not include $z$ in the training set?} Specifically, the change of prediction can be defined as:
\begin{equation}
	\Delta g(u_{t},i_{t},\hat{\theta}_{-z}) \defeq g(u_{t},i_{t},\hat{\theta}_{-z}) - g(u_{t},i_{t},\hat{\theta})
\end{equation}
where $\hat{\theta}_{-z} \defeq \argmin_{\theta} \sum_{z_j \neq z}L(z_j, \theta)$. 
In order to learn $g(u_{t},i_{t},\hat{\theta}_{-z})$ without model retraining, we directly compute $\Delta g(u_{t},i_{t},\hat{\theta}_{-z})$ via influence analysis, as described in Section~\ref{sec:pre:if}. To be specific, the computation of the prediction change involves the following three steps. 
\begin{enumerate}[(i)]
	\item The first step is to measure how upweighting $z$ by an infinitesimal step $\epsilon$ influences the LFM parameters $\hat{\theta}$, i.e., $\frac{{\rm d} \hat{\theta}_{\epsilon,z}}{{\rm d} \epsilon}\big|_{\epsilon=0}$. Recall that $\hat{\theta}_{\epsilon,z}$ is the new parameters learned after upweighting, which can be computed using Equation (\ref{eq:dtepsilon}).
	\item The second step is to measure how upweighting $z$ affects the prediction of LFM at $(u_t,i_t)$ based on the chain rule:
	\begin{equation}
	\label{eq:depsilon}
	\begin{aligned}
	\frac{{\rm d} g(u_t,i_t,\hat{\theta}_{\epsilon,z})}{{\rm d} \epsilon}\Big|_{\epsilon=0}
	&=\nabla_{\theta}g(u_t,i_t,\hat{\theta})^\top \frac{{\rm d} \hat{\theta}_{\epsilon,z}}{{\rm d} \epsilon}\Big|_{\epsilon=0}\\
	&=-\nabla_{\theta}g(u_t,i_t,\hat{\theta})^\top H_{\hat{\theta}}^{-1}\nabla_{\theta}L(z,\hat{\theta})
	\end{aligned}
	\end{equation}
	\item The third step is to approximate $\Delta g(u_{t},i_{t},\hat{\theta}_{-z})$ with the derivative in Equation (\ref{eq:depsilon}) by setting $\epsilon=-\frac{1}{n}$:
	\begin{equation}
	\label{eq:delta_y}
	\Delta g(u_{t},i_{t},\hat{\theta}_{-z})
	\approx \frac{1}{n}\nabla_{\theta}g(u_t,i_t,\hat{\theta})^\top H_{\hat{\theta}}^{-1}\nabla_{\theta}L(z,\hat{\theta})
	\end{equation}
\end{enumerate}
%
%

Based on Equation (\ref{eq:delta_y}), we can obtain two sets of prediction differences $\{\Delta g(u_{t},i_{t},\hat{\theta}_{-z})\mid z \in \CMcal{R}_{u_t}\}$ and $\{\Delta g(u_{t},i_{t},\hat{\theta}_{-z})\mid z \in \CMcal{R}_{i_t}\}$ for the test case ($u_t, i_t$), by examining training points in $\CMcal{R}_{u_t}$ and $\CMcal{R}_{i_t}$, respectively. 

To deliver item-based neighbor-style explanations with the computed prediction differences, we sort the training points $z$ in $\CMcal{R}_{u_t}$ in the descending order of their absolute influence values, i.e., $|\Delta g(u_{t},i_{t},\hat{\theta}_{-z})|$. After that, 
we extract $k$ training points in $\CMcal{R}_{u_t}$ with the largest values of $|\Delta g(u_{t},i_{t},\hat{\theta}_{-z})|$, and treat them as the top-$k$ influential records from user $u_t$ for the rating prediction against item $i_t$, i.e., $k$ item-based explanations as illustrated in Figure~\ref{subfig:item-based}.

Similarly, we can sort the training points in $\CMcal{R}_{i_t}$ and find top-$k$ influential records associated with item $i_t$ to deliver the user-based neighbor-style explanations, as shown in Figure~\ref{subfig:user-based}.

\subsection{Fast Influence Analysis for MF}
\label{sec:FIA}

It is worth noticing that computing $\Delta g(u_{t},i_{t},\hat{\theta}_{-z})$ in Equation (\ref{eq:delta_y})
is expensive due to the existence of the Hessian and its inverse $H_{\hat{\theta}}^{-1}$. Given a training set with $n$ points and an MF model with $p$ parameters in total, the complexity of computing $H_{\hat{\theta}}$ is ${\rm O}(np^2)$ and  reverting $H_{\hat{\theta}}$ needs ${\rm O}(p^3)$ operations.
$p$ reflects model complexity and is determined by the total number of users and items; $n$ can be huge in order to learn better user and item latent representations. To make things worse, for each test case $(u_t, i_t)$, we need to compute $\Delta g$ for all the training data in $\CMcal{R}_t$, recall $\CMcal{R}_t \defeq \CMcal{R}_{u_t}\!\cup\!\CMcal{R}_{i_t}$.


\noindent{\bf Basic influence computation.}
Instead of explicitly computing $H_{\hat{\theta}}^{-1}$, a more efficient way is to compute $\Delta g(u_{t},i_{t},\hat{\theta}_{-z})$ in Equation (\ref{eq:delta_y}) with Hessian-vector products (HVPs) and the iterative algorithm proposed in~\cite{DBLP:conf/icml/KohL17}, which consists of three major steps as follows.
\begin{enumerate}[S1.]
	\item Computing $H_{\hat{\theta}}^{-1} \nabla_{\theta}g(u_t,i_t,\hat{\theta})$. The computation can be transformed into an optimization problem: 
\begin{equation}
	\label{eq:hvp}
		\noindent
	\!H_{\hat{\theta}}^{-1} \nabla_{\theta}g(u_t,i_t,\hat{\theta})\!=\!\argmin_t\{\frac{1}{2}t^\top\! H_{\hat{\theta}}t-\!\nabla_{\theta}g(u_t,i_t,\hat{\theta})^\top\! t\}
	\end{equation} 
	The optimization problem can be solved with conjugate gradients methods, which will  empirically reach convergence within a few iterations~\cite{DBLP:conf/icml/Martens10}.
	Recall that $H_{\hat{\theta}} \defeq \frac{1}{n}\sum_{i=1}^n \nabla_{\theta}^2 L(z_i, \hat{\theta})$, the complexity of this step is ${\rm O}(np)$, which is determined by the computation of $H_{\hat{\theta}}t$~\cite{DBLP:journals/neco/Pearlmutter94}.
	\item Computing $\nabla_{\theta}L(z,\hat{\theta})$. For a training point $z$, getting $\nabla_{\theta}L(z,\hat{\theta})$ requires ${\rm O}(p)$ operations. And since we need to traverse all the training points in $\CMcal{R}_{t}$ to find influential explanations for the rating prediction of ($u_t, i_t$), the complexity of this step is ${\rm O}(p|\CMcal{R}_{t}|)$.
	\item Computing $\Delta g(u_{t},i_{t},\hat{\theta}_{-z})$. Note that $H_{\hat{\theta}}^{-1}$ is symmetric and we can perform this step by combining the results from the previous two steps using Equation (\ref{eq:delta_y}):
	\begin{equation}
	\label{eq:delta_y_adap}
	\Delta g(u_{t},i_{t},\hat{\theta}_{-z})
	= \frac{1}{n}\nabla_{\theta}L(z,\hat{\theta})^\top H_{\hat{\theta}}^{-1}\nabla_{\theta}g(u_t,i_t,\hat{\theta}))
	\end{equation}
	This step needs to perform an inner product with ${\rm O}(p)$ operations for each training point in $\CMcal{R}_{t}$. Hence, the complexity of this step is ${\rm O}(p|\CMcal{R}_{t}|)$.
\end{enumerate}
Let $n'= |\CMcal{R}_{t}|$. Since $n\gg n'$ in practice, the overall complexity of computing $\{\Delta g(u_{t},i_{t},\hat{\theta}_{-z_j})\}_{j=1}^{n'}$ based on the above three steps is ${\rm O}(np)$.


\noindent {\bf Fast influence analysis (FIA).}
Although the basic influence computation with the complexity of ${\rm O}(np)$ is significantly efficient than explicitly calculating $H_{\hat{\theta}}^{-1}$, it still incurs high computation cost over real datasets. According to our experiments, when we employ the aforementioned computation process on the Movielens 1M dataset~\cite{DBLP:journals/tiis/HarperK16}, it takes up to an hour to measure influence of training points on merely one test case, which is obviously unacceptable in nowadays recommender systems. 

To accelerate the influence analysis process, we propose a Fast Influence Analysis algorithm (FIA) for MF based on the following two key observations.
\begin{enumerate}[O1.]
\item For a given test case $(u_t,i_t)$, only a small fraction of MF parameters contribute to the prediction of $y_t$. Specifically, in MF, the prediction of $y_t$ is determined by $\theta_t=\{\mathbf{p}_{u_t}, \mathbf{q}_{i_t}\}$, where $\mathbf{p}_{u_t}$ and $\mathbf{q}_{i_t}$ are the latent vectors for $u_t$ and $i_t$, respectively.
Hence, only the change of parameters in $\theta_t$ affects the prediction of $y_t$.


\item Now that we focus on the analysis of the parameters in $\theta_t$, we only need to measure the influence of training points in $\CMcal{R}_t$ on $\theta_t$. Other training points do not generate gradients on $\theta_t$ and can be ignored during influence computation.
\end{enumerate}
Based on the above observations, we can derive the change of MF's prediction when removing a training point $z$ by:
%
\begin{equation}
\label{eq:MF_delta}
	\Delta g(u_{t},i_{t},\hat{\theta}_{-z}) =
	\frac{1}{n'}\nabla_{\theta_t}g(u_t,i_t,\hat{\theta})^\top H_{\hat{\theta}_t}^{-1}\nabla_{\theta_t}L(z,\hat{\theta})
\end{equation}

Recall that $n'\!=\!|\CMcal{R}_{t}|$.
To better understand the above equation, we zoom into its two parts.  
First, $\nabla_{\theta_t}g(u_t,i_t,\hat{\theta})$ measures how the change of $\theta_t$ affects MF's prediction at $(u_t,i_t)$. We only consider the parameters in $\theta_t$, which is different from the basic influence computation method. 
Second, $\frac{1}{n'}H_{\hat{\theta}_t}^{-1}\nabla_{\theta_t}L(z,\hat{\theta})$ measures how $\theta_t$ changes when removing a training point $z$, and
$H_{\hat{\theta}_t}$ is the Hessian $\frac{1}{n'}\sum_{j=1}^{n'} \nabla_{\theta_t}^2 L(z_j, \hat{\theta})$. 
Here we are able to  
extract a subproblem from the original one, and measure the effects of $\CMcal{R}_t$ on $\theta_t$ since $\nabla_{\theta_t}L(z,\hat{\theta})=\nabla_{\theta_t}^2 L(z, \hat{\theta})=0$ for $z \notin \CMcal{R}_t$.


\noindent {\bf Time complexity of \modelns.}
The computation of Equation (\ref{eq:MF_delta}) can be achieved by the three steps used for calculating Equation (\ref{eq:delta_y_adap}). However, the computation cost involved in Equation (\ref{eq:MF_delta}) is significantly reduced compared with Equation (\ref{eq:delta_y_adap}). Specifically, we analysis the time complexity of each step for computing Equation (\ref{eq:MF_delta}) as follows.
\begin{enumerate}[S1'.]
	\item Computing $H_{\hat{\theta}_t}^{-1} \nabla_{\theta_t} g(u_t,i_t,\hat{\theta})$. According to Equation (\ref{eq:hvp}), the time cost of this step results from the computation of $H_{\hat{\theta}_t}t$, which is ${\rm O}(n'K)$ and $K$ is the dimension of latent vectors. Since now we only need to compute the Hessian of $n'$ points, and the number of parameters $\hat{\theta}_t$ is reduced from $p$ to $2K$.
	\item Computing $\nabla_{\theta_t}L(z,\hat{\theta})$.
	For a training point $z$, computing $\nabla_{\theta_t}L(z,\hat{\theta})$ needs ${\rm O}(K)$ operations. We have to traverse all the training points in $\CMcal{R}_t$ and the complexity of the step becomes ${\rm O}(n'K)$.
	\item Computing $\Delta g(u_{t},i_{t},\hat{\theta}_{-z})$. In this step, we perform an inner product over $H_{\hat{\theta}_t}^{-1} \nabla_{\theta_t} g(u_t,i_t,\hat{\theta})$ and $\nabla_{\theta_t}L(z,\hat{\theta})$. In \modelns, the dimension of the above vectors is $K$, and hence the complexity of traversing $\CMcal{R}_t$ is ${\rm O}(n'K)$.
\end{enumerate}
%
%
%
To sum up, the overall complexity of \model for MF is ${\rm O}(n'K)$, which is greatly reduced compared with ${\rm O}(np)$ cost of the original process, since $n'\ll n$ and $K\ll p$. It is also worth mentioning that both $n'$ and $K$ are typically small and independent of the scale of the training set. 
On the contrary, $n$ and $p$ are proportional to the number of training examples. Thus \model enables us to perform efficient influence analysis for MF even over large datasets.
%

\subsection{Approximate Influence Analysis for NCF}

\noindent {\bf Extending \model to NCF settings.}
The NCF methods are based on the latent factors (i.e., \emph{embeddings}) of users and items, and aim to learn a complex interaction function (e.g., MLP) from training examples unlike performing inner product as MF. 
To adapt \model to NCF settings, we divide the parameters involved in NCF into two parts: $\theta_e$ and $\theta_n$. $\theta_e$ is the latent factors of users and items, $\theta_n$ is the parameters in the interaction function (e.g., weight matrices in MLP). 
Given a test case $(u_t, i_t)$, the rating $y_t$ predicted by NCF is determined by both $\theta_{e}$ and $\theta_{n}$, but the two parts of parameters are affected by training points in different ways. That is, $\theta_{e}$ is optimized by the training points in $\CMcal{R}_t$, where$ \CMcal{R}_t\defeq \CMcal{R}_{u_t} \cup \CMcal{R}_{i_t}$, which is similar to $\theta_t$ in MF, while $\theta_{n}$ is learned using the complete set of $\CMcal{R}$. According to the Taylor expansion of $g(u_t,i_t,\theta_e,\theta_n)$ at $(\hat{\theta}_e,\hat{\theta}_n)$, we have:
\begin{equation}
\label{eq:taylor}
\begin{aligned}
g(u_t,i_t,\theta_e,\theta_n) = g&(u_t,i_t, \hat{\theta}_e,\hat{\theta}_n) + \frac{\partial g}{\partial \theta_e}\Big|_{\theta_e=\hat{\theta}_e}\Delta \theta_e \\
&+ \frac{\partial g}{\partial \theta_n}\Big|_{\theta_n=\hat{\theta}_n}\Delta\theta_n + o(||\Delta \theta||)
\end{aligned} 
\end{equation}
We then compute $\Delta g(u_t,i_t,\hat{\theta}_{-z})$ by dividing it into two parts: $\Delta g(u_t,i_t,\hat{\theta}_{e-z})$ and $\Delta g(u_t,i_t,\hat{\theta}_{n-z})$, and drop the ${\rm o}(||\Delta \theta||)$ term, where $\hat{\theta}_{e-z}$ and $\hat{\theta}_{n-z}$ are respectively the learned $\hat{\theta}_{e}$ and $\hat{\theta}_{n}$ after removing $z$:
\begin{equation}
\label{eq:sum_loss}
\Delta g(u_t,i_t,\hat{\theta}_{-z}) \approx
\Delta g(u_t,i_t,\hat{\theta}_{e-z}) + \Delta g(u_t,i_t,\hat{\theta}_{n-z})
\end{equation}
where $\Delta g(u_t,i_t,\hat{\theta}_{e-z})$ is the change of NCF's prediction due to the changes of the learned embedding when removing a training point $z$. The computation of $\Delta g(u_t,i_t,\hat{\theta}_{e-z})$ is similar to Equation (\ref{eq:MF_delta}):
\begin{equation}
\label{eq:embed_diff}
	\Delta g(u_{t},i_{t},\hat{\theta}_{e-z}) =
	\frac{1}{n'}\nabla_{\theta_e}g(u_t,i_t,\hat{\theta})^\top H_{\hat{\theta}_e}^{-1}\nabla_{\theta_e}L(z,\hat{\theta})
\end{equation}
For the parameters $\theta_n$ involved in the interaction function, we have:
\begin{equation}
\label{eq:nn_loss}
	\Delta g(u_{t},i_{t},\hat{\theta}_{n-z}) =
	\frac{1}{n}\nabla_{\theta_n}g(u_t,i_t,\hat{\theta})^\top H_{\hat{\theta}_n}^{-1}\nabla_{\theta_n}L(z,\hat{\theta})
\end{equation}
Combining Equation (\ref{eq:sum_loss})$-$(\ref{eq:nn_loss}), we can get $\Delta g(u_t,i_t,\hat{\theta}_{-z})$ for NCF recommendation methods. 

\noindent {\bf Time complexity analysis.}
The complexity of influence analysis based on Equation (\ref{eq:embed_diff}) and (\ref{eq:nn_loss}) using \model is similar to that of Equation (\ref{eq:MF_delta}) for MF.
The complexity of computing Equation (\ref{eq:embed_diff}) is ${\rm O}(n'K)$ since the calculation is only based on the parameters of latent factors and the training points in $\CMcal{R}_t$. For Equation (\ref{eq:nn_loss}), the computation cost is ${\rm O}(n|\theta_n|)$,
since the learning of $\theta_n$ relies on all the training points. 
Hence, the total time complexity of \model for NCF is ${\rm O}(n'K+n|\theta_n|)$.

\noindent {\bf FIA for NCF with approximation.}
We observe that in practice, $\Delta g(u_t,i_t,\hat{\theta}_{e-z}) \gg \Delta g(u_t,i_t,\hat{\theta}_{n-z})$, due to the fact that the coefficient $\frac{1}{n'} \gg \frac{1}{n}$, especially for large datasets. For the Movielens 1M dataset, $n'$ is usually a few hundred, while $n$ can be up to one million. This inspires us to compute $\Delta g(u_t,i_t,\hat{\theta}_{-z})$ approximately by dropping the second term in Equation (\ref{eq:sum_loss}) and combining with Equation (\ref{eq:embed_diff}):
\begin{equation}
\Delta g(u_t,i_t,\hat{\theta}_{-z}) \approx
\frac{1}{n'}\nabla_{\theta_e}g(u_t,i_t,\hat{\theta})^\top H_{\hat{\theta}_e}^{-1}\nabla_{\theta_e}L(z,\hat{\theta})
\end{equation}
The intuition for the above approximation is that the effect of removing a training point $z$ on $\theta_e$ is more significant than that on $\theta_n$, since $\theta_n$ is trained on the whole training set instead of the smaller subset $\CMcal{R}_t$.
As a result, the time complexity of \model for NCF methods can be further reduced to ${\rm O}(n'K)$, which is as efficient as \model for MF.

\section{Experiments}

The major contributions of this work are to enforce explainability for LFMs by measuring the influence of training points on the prediction results, and the proposed fast influence analysis methods.
In this section, we conduct experiments to answer the following research questions:
\begin{itemize}
	\item $\mathbf{RQ1}$: How can we prove the correctness of our proposed influence analysis method? Can influence functions be successfully applied to explain the prediction results from LFMs?
	\item $\mathbf{RQ2}$: Can \model compute influence efficiently for LFMs? How does \model perform compared with the basic influence computation process in terms of the efficiency?
	\item $\mathbf{RQ3}$: How do the explanations provided by our methods look like? What insights can we gain from the results of influence analysis for LFMs?
\end{itemize}

\subsection{Experimental Settings}

\subsubsection{Datasets.}

We conduct experiments on two publicly accessible datasets:
\begin{itemize}
	\item \textbf{Yelp}: This is the Yelp Challenge dataset\footnote{https://www.yelp.com/dataset/challenge}, which includes users' ratings on different types of business places (e.g., restaurants, shopping malls, etc).
	\item \textbf{Movielens}: This is the widely used Movielens 1M dataset\footnote{https://grouplens.org/datasets/movielens/1m/}, which contains user ratings on movies.
\end{itemize}
%

In the following experiments, we apply the common preprocessing method~\cite{DBLP:conf/uai/RendleFGS09} to filter out users and items with less than 10 interactions in the datasets. The statistics of the remaining records in two datasets are summarized in Table~\ref{tab:stat}.
\begin{table}[t]
	\centering
	\caption{Statistics of the datasets}\label{tab:stat}
	\vspace{-.1in}
	\resizebox{\linewidth}{!}{
		\begin{tabular}{|c|c|c|c|c|}
			\hline  
			\bf{Dataset}&\bf{Interaction\#}&\bf{User\#}&\bf{Item\#}&\bf{Sparsity}\\
			\hline  
			Yelp&731,671&25,815&25,677&99.89\%\\
			\hline
			Movielens&1,000,209&6,040&3,706&95.53\%\\
			\hline 
		\end{tabular}
	}	
\end{table}

\subsubsection{Parameter Settings.}

We implemented \model using Tensorflow\footnote{https://www.tensorflow.org}.
For each user in the dataset, we randomly held-out one rating as the test set, and used the remaining data for training. We adopted Adam~\cite{adam} to train MF and NCF models, which is a variant of stochastic gradient descent that dynamically tunes the learning rate during training and leads to faster convergence. We set the initial learning rate to 0.001, the batch size to 3000, and the l2 regularization coefficient to 0.001. When computing the influence functions, to avoid negative eigenvalues in Hessian~\cite{DBLP:conf/icml/KohL17}, we add a damping term of $10^{-6}$. All the experiments were conducted on a server with 2 Intel Xeon 1.7GHz CPUs and 2 NVIDIA Tesla K80 GPUs.
 
\subsection{Verification of Correctness (RQ1)}

\begin{figure*}[t]
	\centering
	\subcaptionbox{{\modelns-MF} --- {Yelp} \label{yelp-mf}}
	[.24\linewidth]{\includegraphics[width=.24\linewidth]{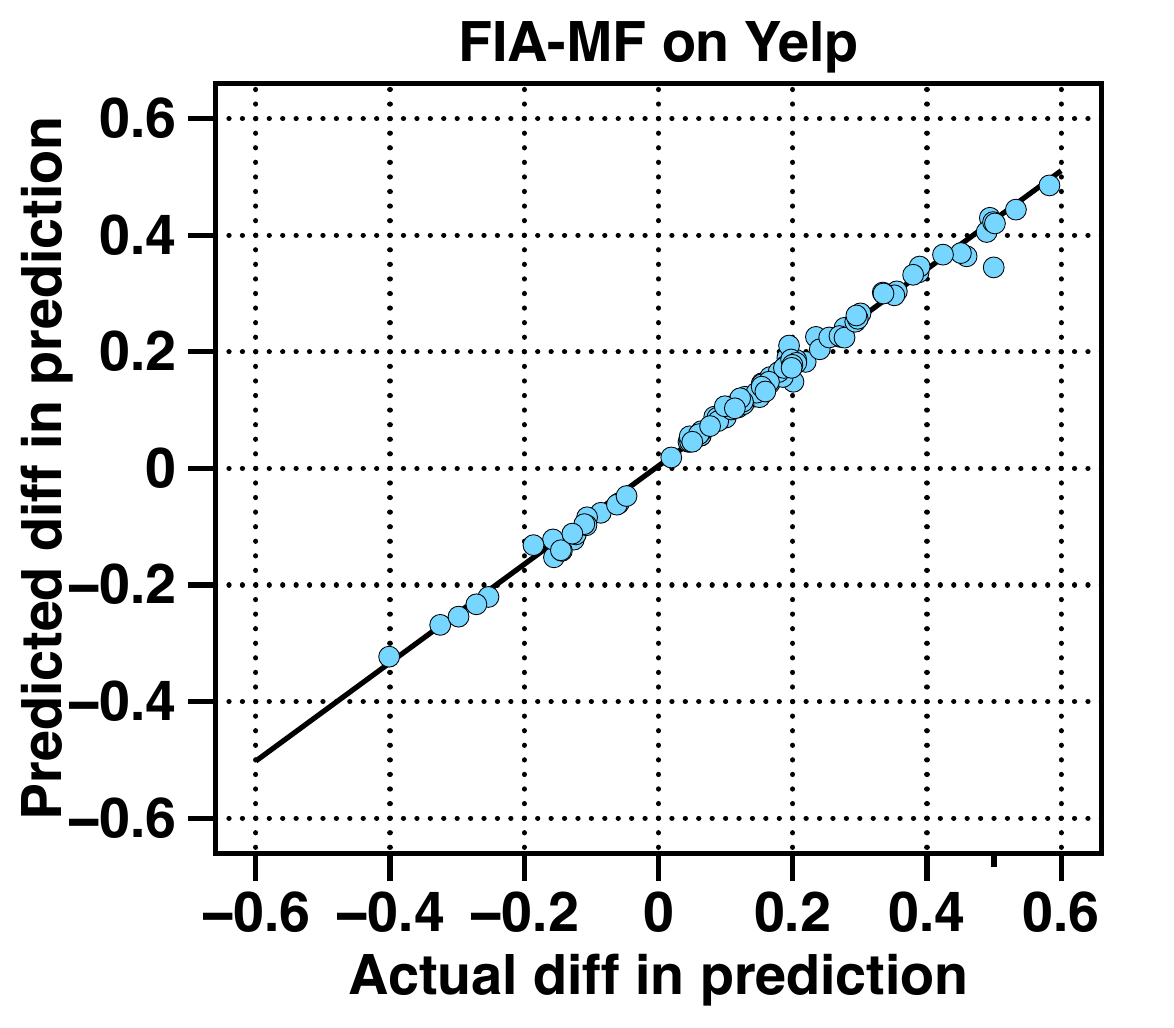}}
	\subcaptionbox{{FIA-MF} --- {Movielens}\label{movie-mf}}
	[.24\linewidth]{\includegraphics[width=.24\linewidth]{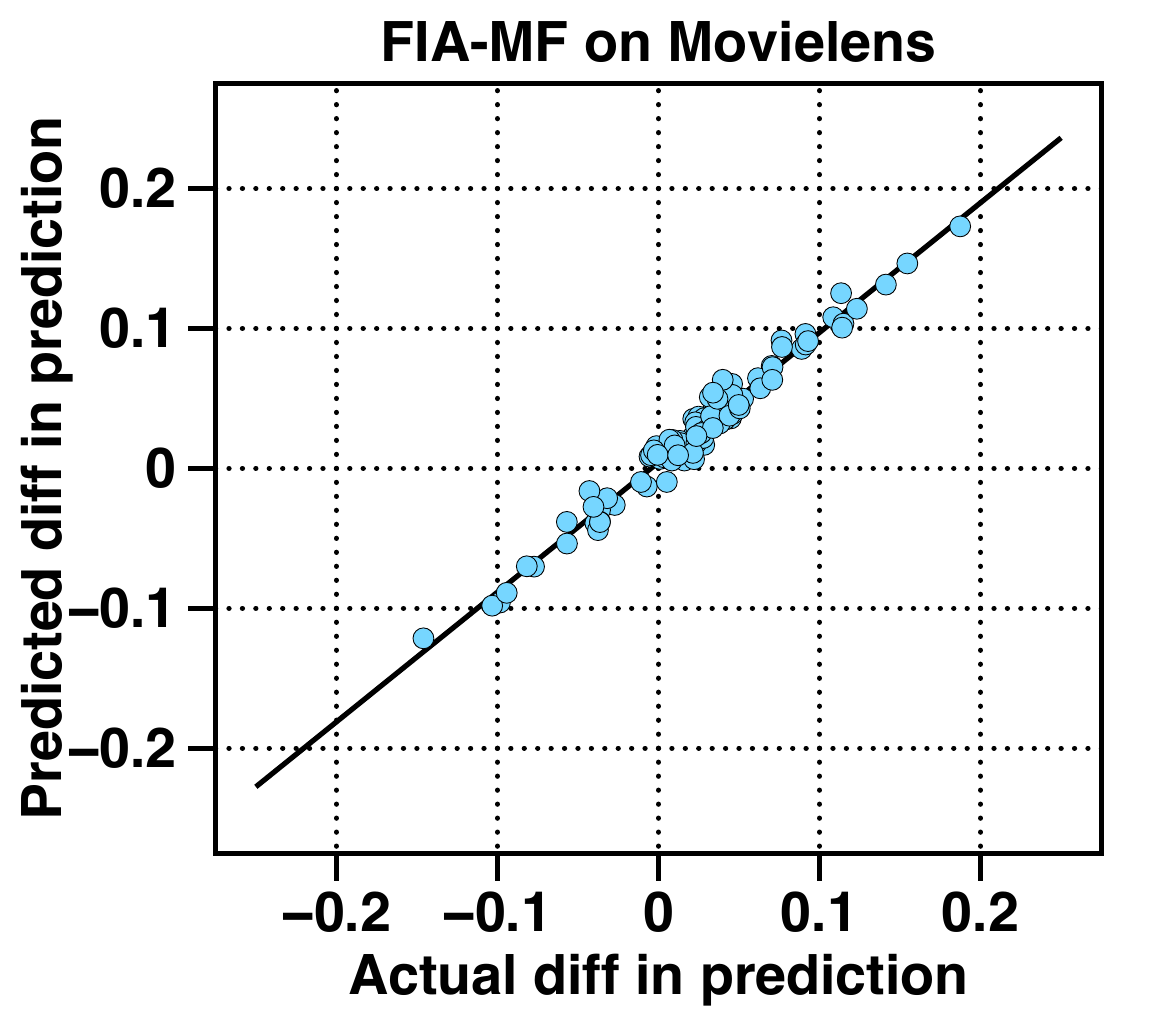}}
	\subcaptionbox{{\modelns-NCF} --- {Yelp}\label{yelp-ncf}}
	[.24\linewidth]{\includegraphics[width=.24\linewidth]{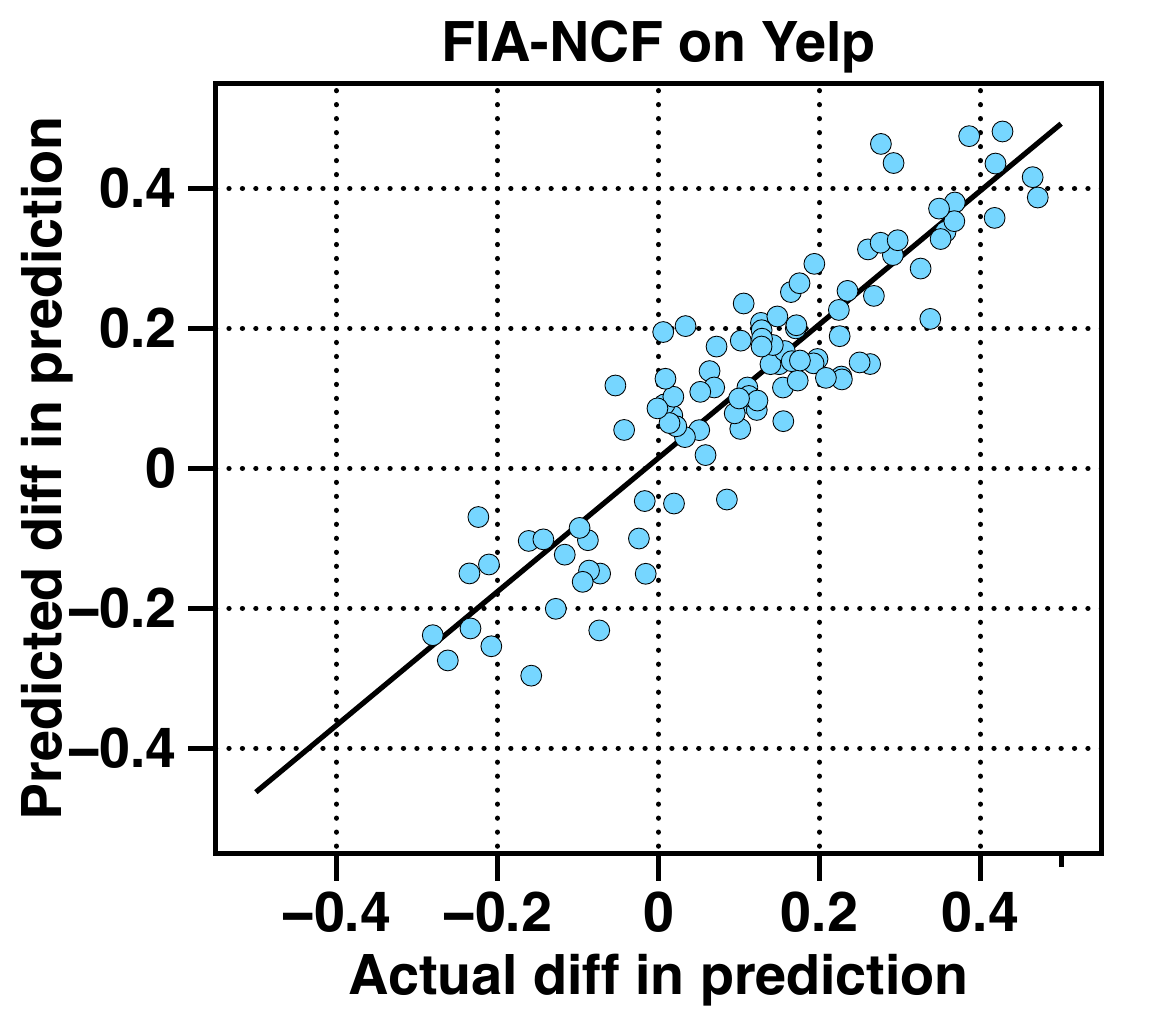}}
	\subcaptionbox{{FIA-NCF} --- {Movielens}\label{movie-ncf}}
	[.24\linewidth]{\includegraphics[width=.24\linewidth]{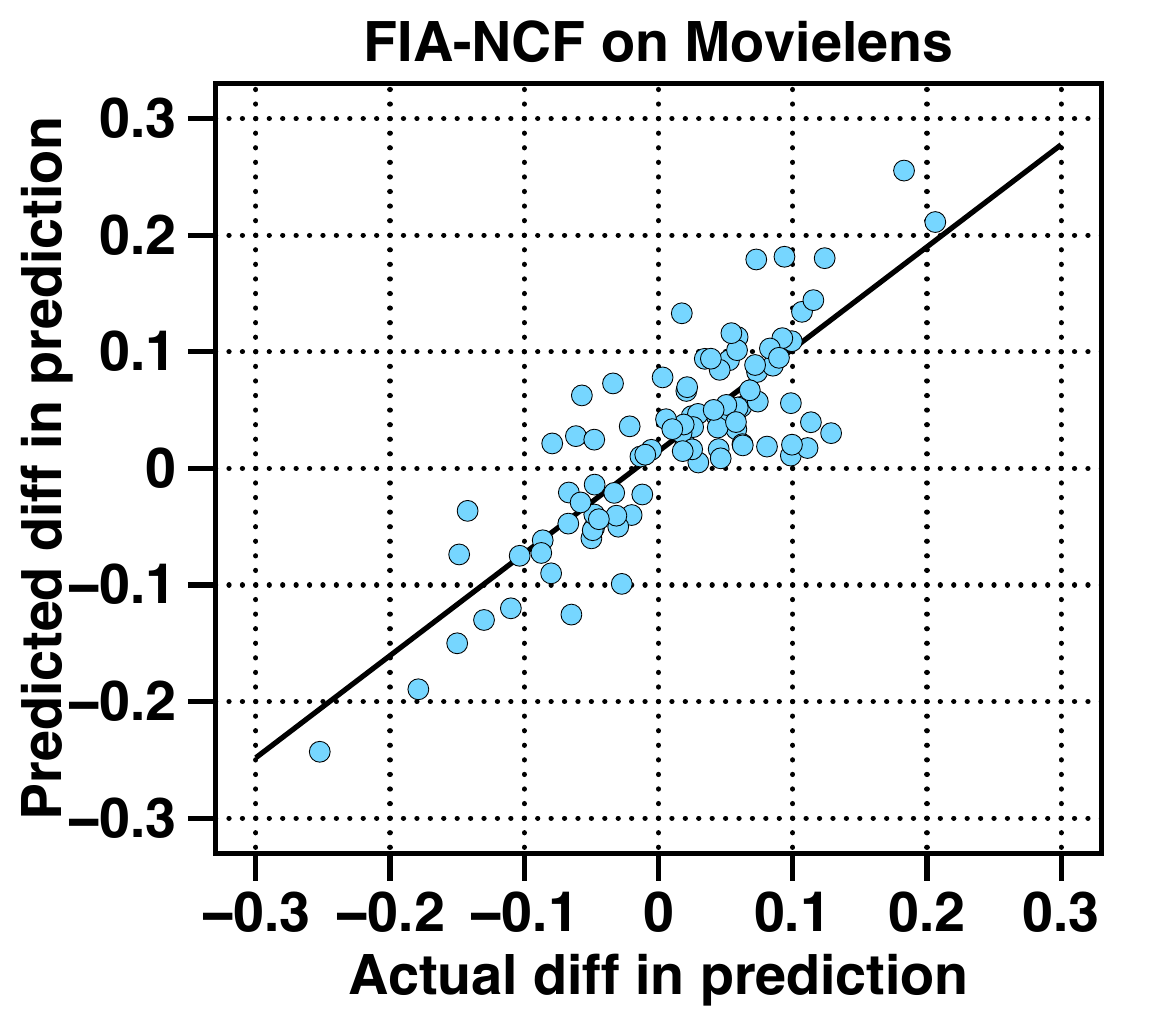}}
\vspace{-0.1in}
	\caption{Comparison between the actual prediction changes and those computed by \model methods for MF and NCF.}
	\label{fig:retrain}
	\vspace{-0.1in}
\end{figure*}

\subsubsection{Evaluation Protocol for Correctness.} 
Given a test case $(u_t, i_t)$, we use \model to compute the prediction changes of  a trained LFM, i.e., $\Delta g(u_t,i_t,\hat{\theta}_{-z})$, by removing a training point $z\in\CMcal{R}_{u_t}\cup\CMcal{R}_{i_t}$ from the whole training set $\CMcal{R}$. To verify the effectiveness of \model on computing $\Delta g(u_t,i_t,\hat{\theta}_{-z})$, we remove the training point $z$ and retrain the model with the remaining points. In this way, we can estimate the true value of $\Delta g$, denoted by $\Delta g_{true}$, and compare it with the estimation result from \modelns. 

In our experiments, we first randomly select 100 test cases $\{(u_{t_i}, i_{t_i})\mid i\in[1,100]\}$ from the test set. For each test case, we apply \model to compute $\{\Delta g(u_{t_i},i_{t_i},\hat{\theta}_{-z})\mid z \in \CMcal{R}_{u_{t_i}}\cup\CMcal{R}_{i_{t_i}}\}$ by considering the influence of all the training points in $\CMcal{R}_{u_{t_i}}\cup\CMcal{R}_{i_{t_i}}$.
Here we only select $\Delta g$ with the largest absolute value and compare it with $\Delta g_{true}$ for correctness verification. 
This is because the true value of prediction change $\Delta g_{true}$ after removing one training point is hard to learn with retraining if it is too small, and is easily overwhelmed by the randomness of retraining process. We also perform retraining multiple times and use the average value as $\Delta g_{true}$ to further alleviate the effects of randomness during retaining.


\subsubsection{Results Analysis.}
We conduct experiments for \modelns-MF and \modelns-NCF on two datasets, where \modelns-NCF is the version with approximation. The results are shown in Figure~\ref{fig:retrain}. First, we can see that the prediction changes computed by \model are highly correlated with their actual values obtained by retraining, which verifies the correctness of \model methods. Specifically, for \modelns-MF, the Pearson correlation coefficient (Pearson's R for brevity) between the computed and actual changes are 0.99 and 0.98 for Yelp and Movielens, respectively. For \modelns-NCF, the Pearson's R between the computed and actual changes are 0.93 and 0.92 for Yelp and Movielens, respectively. The strong correlations between the results from \model and retraining prove that \model methods can effectively approximate the prediction changes without expensive retraining. Besides, it is worth noticing that \model provides better results for MF than NCF on both datasets. This is because in \modelns-NCF, we ignore the effects of a part of model parameters to improve the computational efficiency. This trade-off sacrifices a tiny fraction of accuracy, but we want to emphasize that the approximation method \modelns-NCF can still provide convincing influence analysis according to the results.

\subsection{Study of Computational Efficiency (RQ2)}

\begin{table}[t]
	\centering
	\caption{Running time of \model and IA for MF}\label{tab:tc-mf}
	\vspace{-.1in}
	\resizebox{\linewidth}{!}{
		\begin{tabular}{|p{0.13\linewidth}<{\centering}|p{0.16\linewidth}<{\centering}|p{0.16\linewidth}<{\centering}|p{0.16\linewidth}<{\centering}|p{0.16\linewidth}<{\centering}|}
			\hline 
			&\multicolumn{2}{c|}{\textbf{Yelp}}&\multicolumn{2}{c|}{\textbf{Movielens}}\\
			\hhline{|=|==|==|}  
			\bf{Factors}&\bf{\modelns-MF}&\bf{IA}&\bf{\modelns-MF}&\bf{IA}\\
			\hline
			\bf{8}&0.78s&460s&0.96s&291s\\
			\hline
			\bf{16}&0.75s&500s&1.21s&292s\\
			\hline
			\bf{32}&0.80s&743s&1.59s&456s\\
			\hline
			\bf{64}&0.77s&927s&1.26s&371s\\
			\hline
			\bf{128}&0.95s&1705s&1.24s&167s\\
			\hline
			\bf{256}&0.93s&2242s&1.22s&274s\\
			\hline
		\end{tabular}
	}
\end{table}

\begin{table}[t]
	\centering
	\caption{Running time of \model and IA for NCF}\label{tab:tc-ncf}
	\vspace{-0.1in}
	\resizebox{\linewidth}{!}{
		\begin{tabular}{|p{0.12\linewidth}<{\centering}|p{0.18\linewidth}<{\centering}|p{0.16\linewidth}<{\centering}|p{0.18\linewidth}<{\centering}|p{0.16\linewidth}<{\centering}|}
			\hline 
			&\multicolumn{2}{c|}{\textbf{Yelp}}&\multicolumn{2}{c|}{\textbf{Movielens}}\\
			\hhline{|=|==|==|}  
			\bf{Factors}&\bf{\modelns-NCF}&\bf{IA}&\bf{\modelns-NCF}&\bf{IA}\\
			\hline
			\bf{8}&1.17s&653s&0.89s&405s\\
			\hline
			\bf{16}&1.01s&705s&1.47s&500s\\
			\hline
			\bf{32}&0.97s&1010s&4.01s&601s\\
			\hline
			\bf{64}&0.77s&1350s&4.35s&587s\\
			\hline
			\bf{128}&1.09s&1633s&4.75s&655s\\
			\hline
			\bf{256}&1.38s&2419s&2.41s&712s\\
			\hline
		\end{tabular}
	}
	%
\end{table}

We now empirically evaluate the computational efficiency of \modelns. We measure the time cost of \modelns-MF and \modelns-NCF with IA on the two datasets, where IA is the basic influence computation method that we describe in Section~\ref{sec:FIA}. For each dataset, we randomly select a set of test cases $\{(u_t,i_t)\}$, and we record the average running time for computing the effects of training points on the test cases, i.e., $\{\Delta g(u_{t},i_{t},\hat{\theta}_{-z})\mid z \in \CMcal{R}_{u_{t}}\!\cup\!\CMcal{R}_{i_{t}}\}$.
The running time results with different latent factor dimension settings for MF and NCF are provided in Table~\ref{tab:tc-mf} and Table~\ref{tab:tc-ncf}, respectively. 

From the results, we can see that our \model methods are consistently much more efficient than IA on the two datasets. Specifically, \model runs 135 to 2411 times faster than IA in MF, and achieves a speedup of 138x to 1752x than IA in NCF. Note that the time cost of \model is always at a small scale, i.e., less than 5 seconds, regardless of the dimension of latent factors and the employed datasets. This shows the potential of \model to be applied to real recommendation scenarios. Besides, we can observe that the time cost of each algorithm generally increases with the dimension of latent factors, but with some exceptions. This is caused by the iterative method that we used to solve Equation (\ref{eq:hvp}). That is, the number of iterations until the convergence depends on specific model parameters, which results in certain variance of the running time in practice.

\subsection{Case Study (RQ3)}
\begin{figure}[t]
	\centering
	\includegraphics[width=\linewidth]{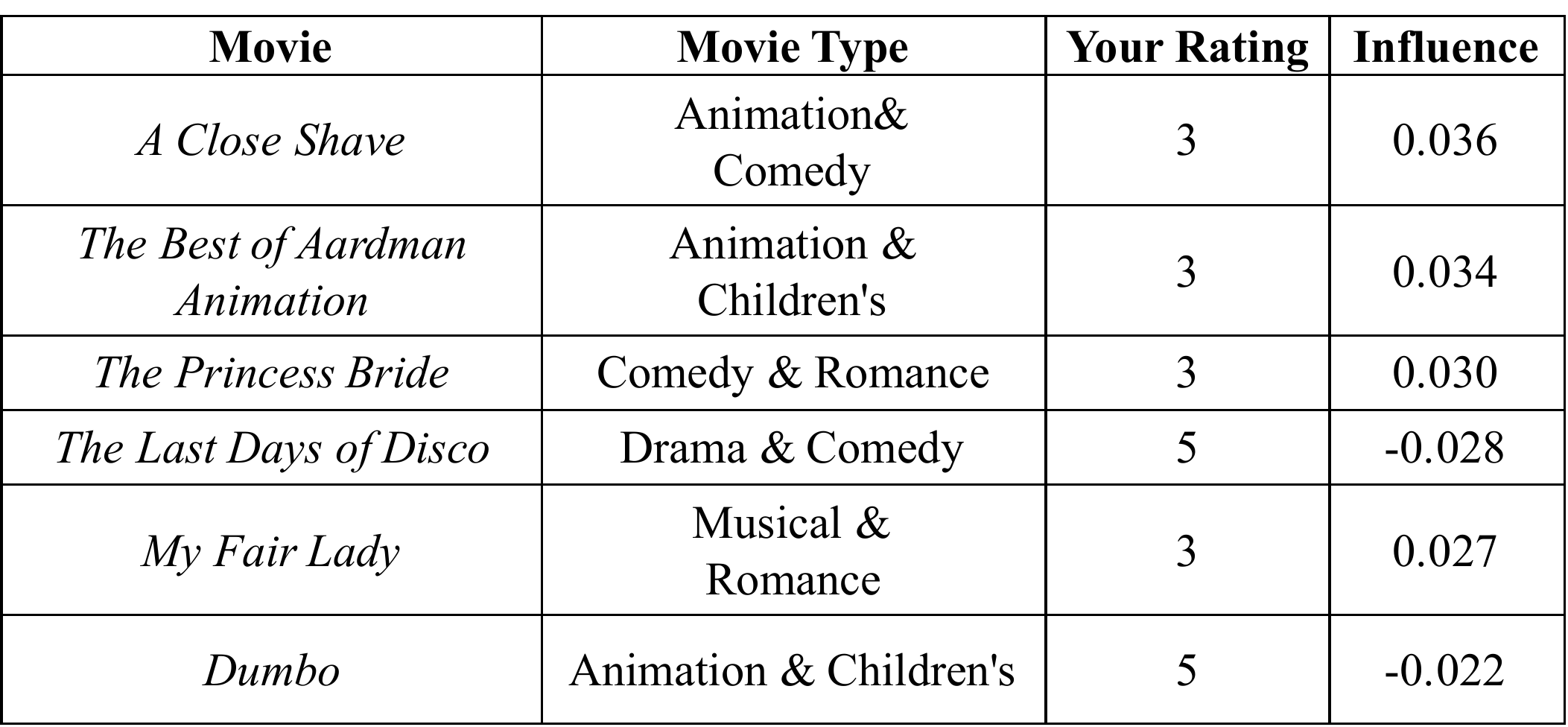}
	\vspace{-0.15in}
	\caption{An example of our item-style explanations.}\label{fig:example}
	\vspace{-0.1in}
\end{figure}

To gain some intuitions on the effectiveness of \model in providing explanations for LFMs, we now give an example for illustration purpose. We use the Movielens dataset and first train a MF model on the dataset until convergence. We then randomly select a test user with 53 historical ratings and predict the user's rating for the movie \emph{The Lion King (1994)} using the trained MF model. In our experiment, the MF method predicts the rating to be $4.04$. Recall that given a trained model and a test case $(u_t, i_t)$, we can compute $\{\Delta g(u_{t},i_{t},\hat{\theta}_{-z})\mid z \in \CMcal{R}_{u_t}\}$ and $\{\Delta g(u_{t},i_{t},\hat{\theta}_{-z})\mid z \in \CMcal{R}_{i_t}\}$ to provide both user-based and item-based neighbor-style explanations for MF. Here we only present item-based explanations, which are typically easier to understand than user-based explanations. More specifically, we compute $\{\Delta g(u_{t},i_{t},\hat{\theta}_{-z})\mid z \in \CMcal{R}_{u_t}\}$ for the test case with \modelns.

The produced explanations are shown in Figure~\ref{fig:example}, where we preserve the top-$6$ influential rating records. We also provide the movie type and the computed influence, to illustrate how the prediction would change when removing the rating from the training set. According to the results, we can explain to the user: ``we predict your rating for \emph{The Lion King (1994)} to be $4.04$, mostly because of your previous ratings on the following 6 items''. Since the type of movie \emph{The Lion King (1994)} belongs to Animation \& Comedy, the explanations in this example are intuitive and convincing. Note that most of movies in the generated list are similar to \emph{The Lion King (1994)} in terms of the movie type. 
 
\begin{figure}[t]
	\centering
	\subcaptionbox{Smooth histogram\label{subfig:sh}}
	[.48\linewidth]{\includegraphics[width=.48\linewidth]{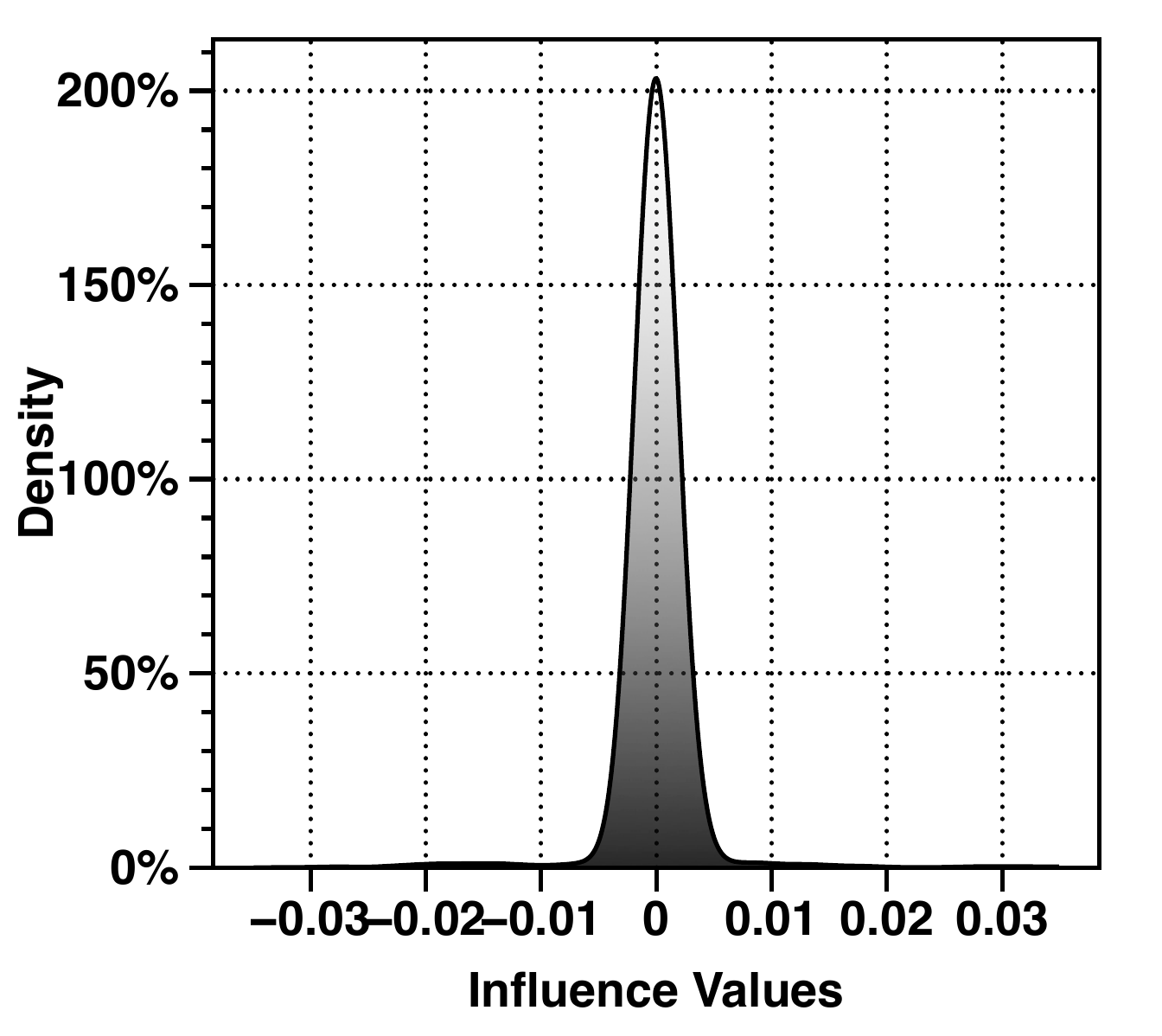}}
	\subcaptionbox{Scatter plot\label{subfig:sp}}
	[.48\linewidth]{\includegraphics[width=.48\linewidth]{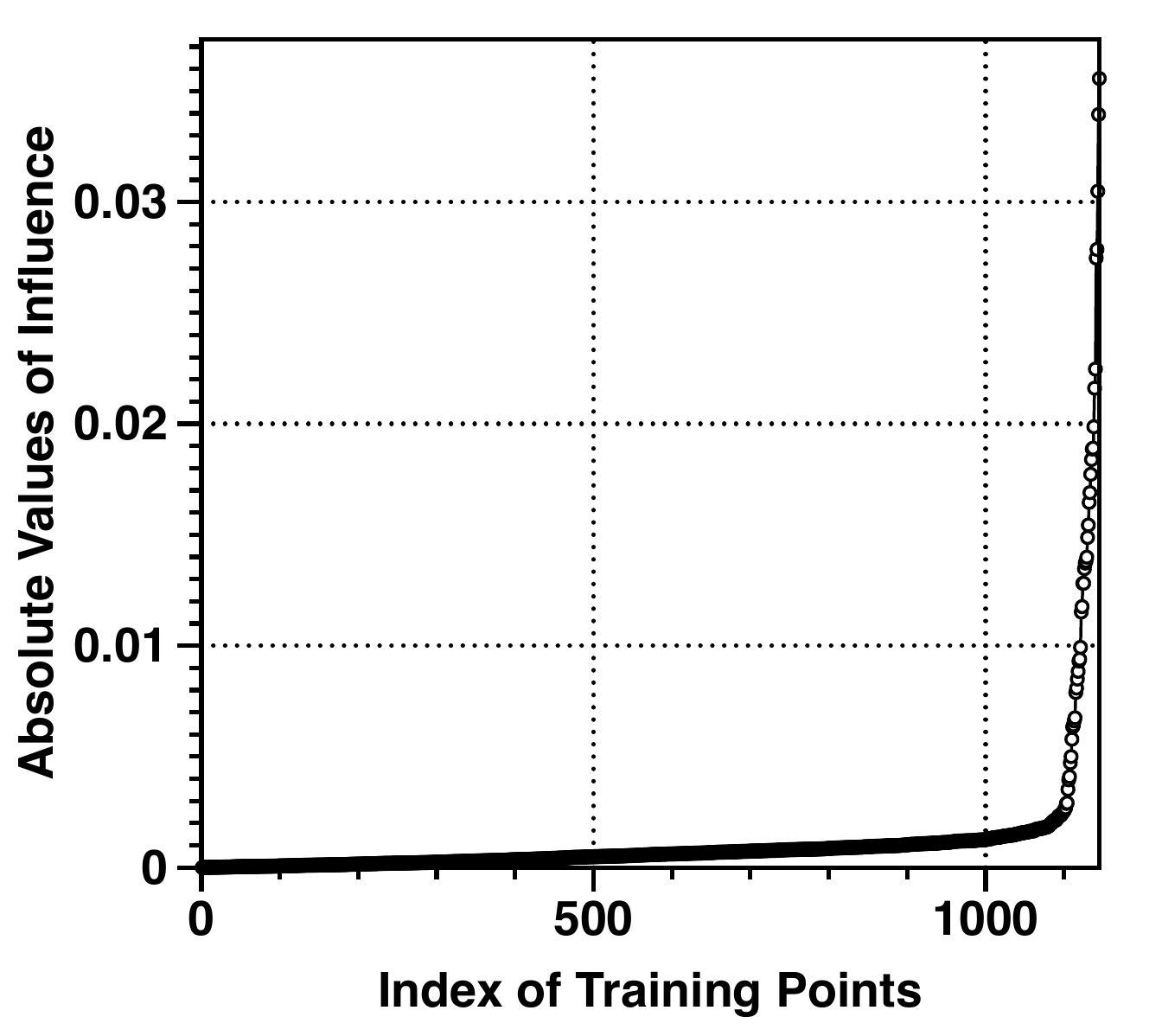}}
	\vspace{-0.1in}
	\caption{Distribution of training points influence.}\label{fig:dis}
	\vspace{-0.1in}
\end{figure}

Besides, to better understand LFM behaviors with influence functions, we draw the distribution of the influence of training points on the test example, which is shown in Figure~\ref{fig:dis}. We focus on the analysis of training points $z \in \CMcal{R}_{u_t}\cup\CMcal{R}_{i_t}$, and Figure~\ref{subfig:sh} is a smooth histogram showing the distribution of influence values. From the figure, we can see that the influence of training points is usually centered around zero and concentrated near zero. Figure~\ref{subfig:sp} is a scatter plot showing the sorted absolute values of the influence scores. We observe that only a small fraction of training points contribute significantly to the MF model's prediction on the test case, which may provide some insights on understanding the security risks of LFMs.



\section{Conclusion}
In this paper, we propose a general method based on influence functions to enforce neighbor-style explanations for LFMs towards explainable recommendation. Our method only leverages the user-item rating matrix without the requirement on auxiliary information. To make influence analysis applicable in real applications, we introduce an efficient computation algorithm for MF. We further extend it to NCF and develop an approximation algorithm to further improve the influence analysis efficiency. Extensive experiments conducted over real-world datasets demonstrate the correctness and efficiency of the proposed method, as well as the usefulness of the provided explanations. 

\newpage
\balance
\bibliography{aaai}

\begin{thebibliography}{}

\bibitem[\protect\citeauthoryear{Abdollahi and
  Nasraoui}{2016a}]{DBLP:conf/www/AbdollahiN16}
Abdollahi, B., and Nasraoui, O.
\newblock 2016a.
\newblock Explainable matrix factorization for collaborative filtering.
\newblock In {\em Proceedings of the 25th International Conference on World
  Wide Web, {WWW} 2016, Montreal, Canada, April 11-15, 2016, Companion Volume},
   5--6.

\bibitem[\protect\citeauthoryear{Abdollahi and
  Nasraoui}{2016b}]{DBLP:journals/corr/AbdollahiN16}
Abdollahi, B., and Nasraoui, O.
\newblock 2016b.
\newblock Explainable restricted boltzmann machines for collaborative
  filtering.
\newblock {\em arXiv preprint arXiv:1606.07129}.

\bibitem[\protect\citeauthoryear{Abdollahi and
  Nasraoui}{2017}]{DBLP:conf/recsys/AbdollahiN17}
Abdollahi, B., and Nasraoui, O.
\newblock 2017.
\newblock Using explainability for constrained matrix factorization.
\newblock In {\em Proceedings of the Eleventh {ACM} Conference on Recommender
  Systems, RecSys 2017, Como, Italy, August 27-31, 2017},  79--83.

\bibitem[\protect\citeauthoryear{Bell and
  Koren}{2007}]{DBLP:journals/sigkdd/BellK07}
Bell, R.~M., and Koren, Y.
\newblock 2007.
\newblock Lessons from the netflix prize challenge.
\newblock {\em {SIGKDD} Explorations} 9(2):75--79.

\bibitem[\protect\citeauthoryear{Bilgic and
  Mooney}{2005}]{bilgic2005explaining}
Bilgic, M., and Mooney, R.~J.
\newblock 2005.
\newblock Explaining recommendations: Satisfaction vs. promotion.
\newblock In {\em Beyond Personalization Workshop, IUI}, volume~5,  153.

\bibitem[\protect\citeauthoryear{Cheng \bgroup et al\mbox.\egroup
  }{2018a}]{DBLP:conf/ijcai/ChengSZH18}
Cheng, W.; Shen, Y.; Zhu, Y.; and Huang, L.
\newblock 2018a.
\newblock {DELF:} {A} dual-embedding based deep latent factor model for
  recommendation.
\newblock In {\em Proceedings of the Twenty-Seventh International Joint
  Conference on Artificial Intelligence, {IJCAI} 2018, July 13-19, 2018,
  Stockholm, Sweden.},  3329--3335.

\bibitem[\protect\citeauthoryear{Cheng \bgroup et al\mbox.\egroup
  }{2018b}]{DBLP:conf/www/ChengDZK18}
Cheng, Z.; Ding, Y.; Zhu, L.; and Kankanhalli, M.~S.
\newblock 2018b.
\newblock Aspect-aware latent factor model: Rating prediction with ratings and
  reviews.
\newblock In {\em Proceedings of the 2018 World Wide Web Conference on World
  Wide Web, {WWW} 2018, Lyon, France, April 23-27, 2018},  639--648.

\bibitem[\protect\citeauthoryear{Cook and
  Weisberg}{1980}]{cook1980characterizations}
Cook, R.~D., and Weisberg, S.
\newblock 1980.
\newblock Characterizations of an empirical influence function for detecting
  influential cases in regression.
\newblock {\em Technometrics} 22(4):495--508.

\bibitem[\protect\citeauthoryear{Cook and Weisberg}{1982}]{cook1982residuals}
Cook, R.~D., and Weisberg, S.
\newblock 1982.
\newblock {\em Residuals and influence in regression}.
\newblock New York: Chapman and Hall.

\bibitem[\protect\citeauthoryear{Harper and
  Konstan}{2016}]{DBLP:journals/tiis/HarperK16}
Harper, F.~M., and Konstan, J.~A.
\newblock 2016.
\newblock The movielens datasets: History and context.
\newblock {\em TiiS} 5(4):19:1--19:19.

\bibitem[\protect\citeauthoryear{He \bgroup et al\mbox.\egroup
  }{2017}]{DBLP:conf/www/HeLZNHC17}
He, X.; Liao, L.; Zhang, H.; Nie, L.; Hu, X.; and Chua, T.
\newblock 2017.
\newblock Neural collaborative filtering.
\newblock In {\em Proceedings of the 26th International Conference on World
  Wide Web, {WWW} 2017, Perth, Australia, April 3-7, 2017},  173--182.

\bibitem[\protect\citeauthoryear{He \bgroup et al\mbox.\egroup
  }{2018}]{DBLP:conf/ijcai/0001DWTTC18}
He, X.; Du, X.; Wang, X.; Tian, F.; Tang, J.; and Chua, T.
\newblock 2018.
\newblock Outer product-based neural collaborative filtering.
\newblock In {\em Proceedings of the Twenty-Seventh International Joint
  Conference on Artificial Intelligence, {IJCAI} 2018, July 13-19, 2018,
  Stockholm, Sweden.},  2227--2233.

\bibitem[\protect\citeauthoryear{Kingma and Ba}{2014}]{adam}
Kingma, D., and Ba, J.
\newblock 2014.
\newblock Adam: A method for stochastic optimization.
\newblock {\em arXiv preprint arXiv:1412.6980}.

\bibitem[\protect\citeauthoryear{Koh and Liang}{2017}]{DBLP:conf/icml/KohL17}
Koh, P.~W., and Liang, P.
\newblock 2017.
\newblock Understanding black-box predictions via influence functions.
\newblock In {\em Proceedings of the 34th International Conference on Machine
  Learning, {ICML} 2017, Sydney, NSW, Australia, 6-11 August 2017},
  1885--1894.

\bibitem[\protect\citeauthoryear{Lee and Seung}{2000}]{DBLP:conf/nips/LeeS00}
Lee, D.~D., and Seung, H.~S.
\newblock 2000.
\newblock Algorithms for non-negative matrix factorization.
\newblock In {\em Advances in Neural Information Processing Systems 13, Papers
  from Neural Information Processing Systems {(NIPS)} 2000, Denver, CO, {USA}},
   556--562.

\bibitem[\protect\citeauthoryear{Martens}{2010}]{DBLP:conf/icml/Martens10}
Martens, J.
\newblock 2010.
\newblock Deep learning via hessian-free optimization.
\newblock In {\em Proceedings of the 27th International Conference on Machine
  Learning (ICML-10), June 21-24, 2010, Haifa, Israel},  735--742.

\bibitem[\protect\citeauthoryear{Pearlmutter}{1994}]{DBLP:journals/neco/Pearlmutter94}
Pearlmutter, B.~A.
\newblock 1994.
\newblock Fast exact multiplication by the hessian.
\newblock {\em Neural Computation} 6(1):147--160.

\bibitem[\protect\citeauthoryear{Rendle \bgroup et al\mbox.\egroup
  }{2009}]{DBLP:conf/uai/RendleFGS09}
Rendle, S.; Freudenthaler, C.; Gantner, Z.; and Schmidt{-}Thieme, L.
\newblock 2009.
\newblock {BPR:} bayesian personalized ranking from implicit feedback.
\newblock In {\em {UAI} 2009, Proceedings of the Twenty-Fifth Conference on
  Uncertainty in Artificial Intelligence, Montreal, QC, Canada, June 18-21,
  2009},  452--461.

\bibitem[\protect\citeauthoryear{Resnick \bgroup et al\mbox.\egroup
  }{1994}]{DBLP:conf/cscw/ResnickISBR94}
Resnick, P.; Iacovou, N.; Suchak, M.; Bergstrom, P.; and Riedl, J.
\newblock 1994.
\newblock Grouplens: An open architecture for collaborative filtering of
  netnews.
\newblock In {\em {CSCW} '94, Proceedings of the Conference on Computer
  Supported Cooperative Work, Chapel Hill, NC, USA, October 22-26, 1994},
  175--186.

\bibitem[\protect\citeauthoryear{Ricci \bgroup et al\mbox.\egroup
  }{2011}]{DBLP:reference/rsh/2011}
Ricci, F.; Rokach, L.; Shapira, B.; and Kantor, P.~B., eds.
\newblock 2011.
\newblock {\em Recommender Systems Handbook}.
\newblock Springer.

\bibitem[\protect\citeauthoryear{Sarwar \bgroup et al\mbox.\egroup
  }{2001}]{DBLP:conf/www/SarwarKKR01}
Sarwar, B.~M.; Karypis, G.; Konstan, J.~A.; and Riedl, J.
\newblock 2001.
\newblock Item-based collaborative filtering recommendation algorithms.
\newblock In {\em Proceedings of the Tenth International World Wide Web
  Conference, {WWW} 10, Hong Kong, China, May 1-5, 2001},  285--295.

\bibitem[\protect\citeauthoryear{Wang \bgroup et al\mbox.\egroup
  }{2018}]{DBLP:conf/sigir/WangWJY18}
Wang, N.; Wang, H.; Jia, Y.; and Yin, Y.
\newblock 2018.
\newblock Explainable recommendation via multi-task learning in opinionated
  text data.
\newblock In {\em The 41st International {ACM} {SIGIR} Conference on Research
  {\&} Development in Information Retrieval, {SIGIR} 2018, Ann Arbor, MI, USA,
  July 08-12, 2018},  165--174.

\bibitem[\protect\citeauthoryear{Zhang and
  Chen}{2018}]{DBLP:journals/corr/abs-1804-11192}
Zhang, Y., and Chen, X.
\newblock 2018.
\newblock Explainable recommendation: {A} survey and new perspectives.
\newblock {\em CoRR} abs/1804.11192.

\bibitem[\protect\citeauthoryear{Zhang \bgroup et al\mbox.\egroup
  }{2014}]{DBLP:conf/sigir/ZhangL0ZLM14}
Zhang, Y.; Lai, G.; Zhang, M.; Zhang, Y.; Liu, Y.; and Ma, S.
\newblock 2014.
\newblock Explicit factor models for explainable recommendation based on
  phrase-level sentiment analysis.
\newblock In {\em The 37th International {ACM} {SIGIR} Conference on Research
  and Development in Information Retrieval, {SIGIR} '14, Gold Coast , QLD,
  Australia - July 06 - 11, 2014},  83--92.

\end{thebibliography}
\bibliographystyle{aaai}
\end{document}